\documentclass[10pt,journal,compsoc]{IEEEtran}

\usepackage[OT1]{fontenc} 
\usepackage{epsfig}
\usepackage{graphicx}
\usepackage{amsmath}
\usepackage{amssymb}
\usepackage{algorithm}
\usepackage{algorithmic}
\usepackage[pagebackref=true,breaklinks=true,citecolor=blue,linkcolor=blue,urlcolor=blue,colorlinks,bookmarks=false]{hyperref}

\newcommand{\tabincell}[2]{\begin{tabular}{@{}#1@{}}#2\end{tabular}}

\newcommand{\etal}{\textit{et al}.}
\newcommand{\ie}{\textit{i}.\textit{e}.,}
\newcommand{\eg}{\textit{e}.\textit{g}.,}
\newcommand{\vs}{\textit{v}.\textit{s}.}

\usepackage{ifpdf}
\usepackage{graphicx} %%Grafiken und normales LaTeX
\usepackage{diagbox}
\usepackage{multirow}
\usepackage{color}
\usepackage{subfigure}
\usepackage{enumitem}
\usepackage{setspace}
\usepackage{booktabs}
\usepackage{makecell}
\usepackage[numbers,sort&compress]{natbib}
\usepackage[dvipsnames]{xcolor}

\usepackage{titlesec}
\titleclass{\subsubsubsection}{straight}[\subsection]

\newcounter{subsubsubsection}[subsubsection]
\renewcommand\thesubsubsubsection{\thesubsubsection.\arabic{subsubsubsection}}

\titleformat{\subsubsubsection}
  {\normalfont\normalsize\bfseries}{\thesubsubsubsection}{1em}{}
\titlespacing*{\subsubsubsection}
{0pt}{3.25ex plus 1ex minus .2ex}{1.5ex plus .2ex}

\makeatletter
\renewcommand\paragraph{\@startsection{paragraph}{5}{\z@}%
  {3.25ex \@plus1ex \@minus.2ex}%
  {-1em}%
  {\normalfont\normalsize\bfseries}}
\renewcommand\subparagraph{\@startsection{subparagraph}{6}{\parindent}%
  {3.25ex \@plus1ex \@minus .2ex}%
  {-1em}%
  {\normalfont\normalsize\bfseries}}
\def\toclevel@subsubsubsection{4}
\def\toclevel@paragraph{5}
\def\toclevel@paragraph{6}
\def\l@subsubsubsection{\@dottedtocline{4}{7em}{4em}}
\def\l@paragraph{\@dottedtocline{5}{10em}{5em}}
\def\l@subparagraph{\@dottedtocline{6}{14em}{6em}}
\makeatother

\begin{document}
%
% paper title
% Titles are generally capitalized except for words such as a, an, and, as,
% at, but, by, for, in, nor, of, on, or, the, to and up, which are usually
% not capitalized unless they are the first or last word of the title.
% Linebreaks \\ can be used within to get better formatting as desired.
% Do not put math or special symbols in the title.
\title{Automatically Discovering Novel Visual Categories with Adaptive Prototype Learning}
%
%
% author names and IEEE memberships
% note positions of commas and nonbreaking spaces ( ~ ) LaTeX will not break
% a structure at a ~ so this keeps an author's name from being broken across
% two lines.
% use \thanks{} to gain access to the first footnote area
% a separate \thanks must be used for each paragraph as LaTeX2e's \thanks
% was not built to handle multiple paragraphs
%
\author{Lu Zhang$^\dag$,~Lu Qi$^\dag$,~Xu Yang,~Hong Qiao, ~Ming-Hsuan Yang, ~Zhiyong Liu
% <-this % stops a space
\IEEEcompsocitemizethanks{\IEEEcompsocthanksitem Lu Zhang, Xu Yang, Hong Qiao and Zhiyong Liu are with State Key Laboratory of Management and Control for Complex Systems, Institute of Automation, Chinese Academy of Sciences, China, and also with the School of Artiﬁcial Intelligence, University of Chinese Academy of Sciences, China. Xu Yang and Zhiyong Liu are also with Center for Excellence in Brain Science and Intelligence Technology, Chinese Academy of Sciences, China\protect\\
Lu Qi and Ming-Hsuan Yang are with University of California at Merced, USA\\
$^\dag$ Both authors contribution equally to this work.\protect\\
% note need leading \protect in front of \\ to get a newline within \thanks as
% \\ is fragile and will error, could use \hfil\break instead.
%Corresponding author: .
%\IEEEcompsocthanksitem J. Doe and J. Doe are with Anonymous University.
}% <-this % stops an unwanted space
%\thanks{Manuscript received April 19, 2005; revised August 26, 2015.}
}

\IEEEtitleabstractindextext{%
\begin{abstract}
This paper tackles the problem of novel category discovery (NCD), which aims to discriminate unknown categories in large-scale image collections. 
The NCD task is challenging due to the closeness to the real world scenarios, where we have only encountered some partial classes and images. 
Unlike other works on the NCD, we leverage the prototypes to emphasize the importance of category discrimination and alleviate the issue with missing annotations of novel classes. 
Concretely, we propose a novel adaptive prototype learning method consisting of two main stages: prototypical representation learning and prototypical self-training. 
In the first stage, we obtain a robust feature extractor, which could serve for all images with base and novel categories. 
This ability of instance and category discrimination of the  feature extractor is boosted 
by self-supervised learning and adaptive prototypes.
In the second stage, we utilize the prototypes again to rectify offline pseudo labels and train a final parametric classifier for category clustering.
We conduct extensive experiments on four benchmark datasets, and demonstrate the effectiveness and robustness of the proposed method with the state-of-the-art performance.
The source code and trained models will be made available at this \href{https://github.com/dvlab-research/Entity}{github} site.
\end{abstract}

% Note that keywords are not normally used for peerreview papers.
\begin{IEEEkeywords}
novel category discovery, image recognition, transfer learning.
\end{IEEEkeywords}}

% make the title area
\maketitle

\IEEEdisplaynontitleabstractindextext
\IEEEpeerreviewmaketitle

\IEEEraisesectionheading{\section{Introduction}}
\IEEEPARstart{R}{ecently,} various computer vision tasks such as image classification~\cite{simonyan2014very, he2016deep} and face recognition~\cite{parkhi2015deep, liu2017sphereface} have obtained significant advances driven by deep learning. 
With the help of large-scale datasets, \eg~ImageNet~\cite{deng2009imagenet} and IBUG-300W~\cite{sagonas2013300}, the trained models of those tasks manifest state-of-the-art recognition ability in new images. 
However, those tasks are usually in the close-set, requiring classifying images to limited pre-defined categories. 
It is intrinsically difficult for trained models to expand the learned knowledge to novel concepts~\cite{orhan2020self, krishnaswamy2022exploiting} as human beings can effortlessly achieve.
For example, young children can discover novel shapes and animals \cite{bomba1983nature, quinn1993evidence} (\eg~triangle and bird) and  differentiate them based on other seen classes \cite{colung2003emergence, han2019learning} 
(\eg~circle and dog). 
This is an innate capability of humans but a great challenge for deep learning models~\cite{han2019learning, serre2019deep}.
Making deep models accommodate to the real world has drawn increasing attention in the vision community.

\begin{figure}[!t]
\centering
\includegraphics[width=3.2in]{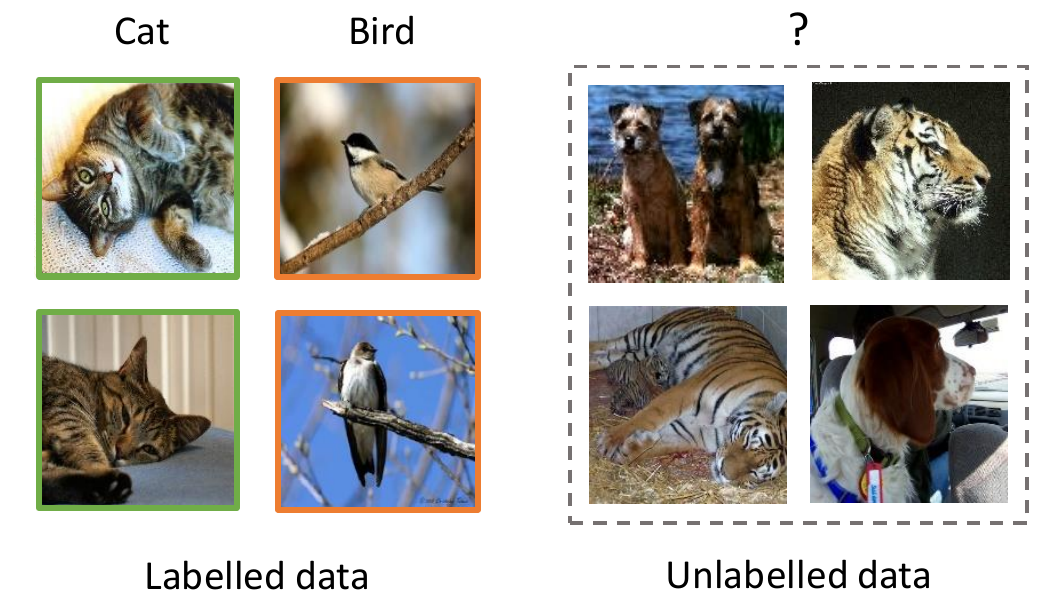}
\caption{Illustration of the task of novel category discovery. Given limited labelled images of some known categories, the model needs to automatically separate unlabelled images of novel categories, thus being able to recognize both old and novel categories in testing data. Images are from the ImageNet \cite{deng2009imagenet} dataset.}
\label{figure_intro_sketch}
\end{figure}

The task of \textit{novel category discovery} (NCD) has been proposed to solve the open-world problem in recent years. 
Given some labelled data of partial categories, it aims at partitioning unlabelled data into some semantic clusters, which we could regard as anonymous classes. 
These clusters are open-world without the limitation to the predefined categories. 
As shown in Figure~\ref{figure_intro_sketch}, the NCD task is closer to the real world scenarios, where we can access enormous data but only very few are annotated with limited categories. 
Thus, the NCD task should require supervised classification and unsupervised clustering methods. 
Inspired by methods developed for learning to cluster~\cite{hsu2018multi, hsu2018learning} that transfer binary labels from labelled to unlabelled data, recent NCD approaches~\cite{han2019learning, han2019automatically, han2021autonovel, jia2021joint} usually utilize pair-wise similarity shared by all categories to produce pseudo-binary labels.
This training pipeline could generate robust instance discrimination that helps separate the novel visual concepts. 
However, we show that both instance and category discrimination are essential to discovering the novel categories in the NCD task. 
We note that the ability of category discrimination is more relevant to the NCD task and has been proven crucial in traditional classification tasks. 
In this work, we aim to boost the model ability of both instance and category discrimination for the NCD task.

We can easily use a more robust self-supervised method to improve instance discrimination. 
For category discrimination, the standard strategy is cross-entropy loss. 
However, adopting a similar strategy is not trivial in the NCD task due to lacking novel class labels. 
Instead, we propose utilizing the adaptive prototypes to encode novel category information, where each prototype represents a class. 
During training, the class prototypes could be dynamically updated by adopting more discriminative instance features with momentum. 
Specifically, our method has two stages, including prototypical representation learning (stage \uppercase\expandafter{\romannumeral1}) and prototypical self-training (stage \uppercase\expandafter{\romannumeral2}). 
%
%The stage \uppercase\expandafter{\romannumeral1} is used to provide the robust image feature with the classifier in stage \uppercase\expandafter{\romannumeral2}. 
%
%In the following, we summarize how to utilize the prototypes in both stages.

%MH: Ideally, you should reduce the descriptions of stage I and II to a few sentences (\eg 50% of the sentences that you have so far) as you mainly want to give an overview of these stages in the first section. I remove a few sentences to make it more compact. 
The aim of stage \uppercase\expandafter{\romannumeral1} is to obtain a robust feature extractor to serve all images, irrespective of base or novel categories.
We achieve it by the self-supervised learning method DINO \cite{caron2021emerging} and the module of online prototype learning (OPL).
On the one hand, DINO is a more robust self-supervised learning method without requiring negative samples. 
Meanwhile, we present a new data augmentation strategy named restricted rotation for multi-view construction of symbolic data (\eg~shape and character). 
On the other hand, the OPL is the critical part of stage I to excavate the inherent category discrimination ability. 
It can maintain adaptive prototypes for novel categories, allowing prototype online updates and then assigning class-level pseudo labels on-the-fly. 
%
% Thus, we can utilize standard cross-entropy loss for both base and novel categories. 
%
% To better push away prototypes with semantically unrelated classes, the pair-wise angular separation is adopted to distribute adaptive prototypes separately. 

In stage~\uppercase\expandafter{\romannumeral2}, we empirically find that online pseudo labels generated in stage~\uppercase\expandafter{\romannumeral1} are less effective (\ie~they are unreliable for training a classifier over the whole dataset). 
As such, in stage~\uppercase\expandafter{\romannumeral2} we retrain a parametric classifier for base and novel categories with three main components: pseudo labelling, prototypical pseudo label rectification, and joint optimization. %we design the prototypical self-training in stage~\uppercase\expandafter{\romannumeral2} to conduct the offline pseudo label rectification for classification. 
The pseudo labelling leverages the discriminative feature extracted by the model trained in stage \uppercase\expandafter{\romannumeral1} and then generates pseudo labels by clustering. 
We then reuse the angular similarity of prototypes to rectify pseudo labels further. 
Finally, we optimize the loss of base data with human labels and novel data with offline pseudo labels. 
The stage~\uppercase\expandafter{\romannumeral2} is optionally iterated to refine the classifier decision boundary and improve recognition accuracy.

Overall, the prototypical representation learning method in stage~\uppercase\expandafter{\romannumeral1} builds a strong feature extractor for non-parametric classification via clustering.
In stage~\uppercase\expandafter{\romannumeral2}, a parametric classifier is trained by prototypical self-training. 
The proposed prototypical learning method facilitates improving the discrimination ability with online and offline pseudo labels. 
To the best of our knowledge, this is the first approach to focus more on category discrimination and effectively make use of prototypes in the NCD task. 
Extensive experiments on benchmark datasets including CIFAR10~\cite{krizhevsky2009learning}, CIFAR100~\cite{krizhevsky2009learning}, OmniGlot~\cite{lake2015human}, and ImageNet~\cite{deng2009imagenet} demonstrate the effectiveness of our method in different settings. 
For novel category discovery, we achieve state-of-the-art performance on the unlabelled set of all datasets.
In addition, our method generalizes better than 
existing schemes in the testing set. 
With the labelled data and ``pseudo-labelled'' unlabelled data, our method can recognize new categories without forgetting the old ones.

\section{Related Work}
\label{S2}
In this section, we first review semi-supervised and self-supervised learning methods due to   close relevance to the usage of labelled and unlabelled data. 
Then, we introduce some methods related to transfer clustering, which motivates our designed two stages and core module of prototypical learning. 

\subsection{Semi-supervised Learning}
Semi-supervised learning is typically used to train a model with a small amount of labelled data and a large amount of unlabelled data. 
Its main challenge is effectively leveraging unlabelled data to improve the model performance.

In the era of deep learning, semi-supervised learning methods can be broadly categorized as: consistency regularization~\cite{bachman2014learning, sajjadi2016regularization, laine2016temporal, tarvainen2017mean} and self-training~\cite{rosenberg2005semi, xie2020self} (\ie~pseudo-labelling~\cite{lee2013pseudo,rizve2020defense,qi2021casp}). 
Consistency regularization methods assume that the model should be less sensitive to the different perturbations imposed on the inputs.
Thus, the model predictions for the unlabelled data can be utilized as artificial labels to enforce consistency. 
Self-training methods first train the model on the labelled data, and then utilize it to generate pseudo labels for the unlabelled data. 
This pseudo-labelling process may iterate to produce better results. 
The idea behind self-training is direct and can be traced back to decades ago~\cite{mclachlan1975iterative, scudder1965probability} before the emergence of deep learning. 
There are also some mixed ideas between consistency regularization and self-training, such as FixMatch~\cite{sohn2020fixmatch} and ISMT~\cite{yang2021interactive}. 
As the intermediate zone of unsupervised and supervised learning, semi-supervised learning has recently witnessed success in combining self-supervised pre-training and  self-training~\cite{chen2020big, zoph2020rethinking, liu2021self}.

In general, the labelled and unlabelled data in semi-supervised learning contain the same object categories. 
However, unlike semi-supervised learning, the NCD task requires recognizing unlabelled novel categories that are not observed in the labelled data. 
Thus, we need some potential classes by unsupervised clustering before applying appropriate semi-supervised methods.

\begin{figure*}[!t]
\centering
\includegraphics[width=7in]{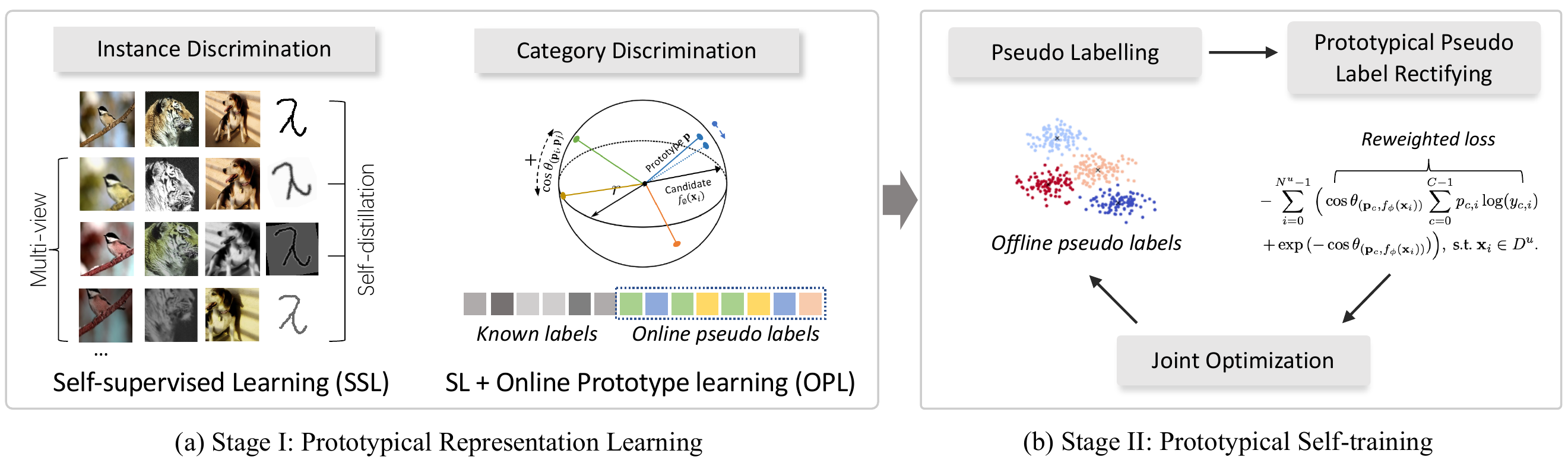}
\caption{Overview of the proposed framework. It contains two training stages: prototypical representation learning (stage I) and prototypical self-training (stage II). First, stage I obtains a robust feature extractor, which could serve for all images with base and novel categories. This feature extractor is boosted in the ability of instance and category discrimination by self-supervised learning and adaptive prototypes. After that, stage II utilizes the prototypes again to 
rectify offline pseudo labels and then train a final parametric classifier for category clustering.}
\label{figure_overview}
\end{figure*}

\subsection{Self-supervised Learning}
Self-supervised learning has recently achieved significant success in natural language and computer vision without requiring expensive target labels. 
The core of self-supervised learning is designing some pretext tasks to obtain better representations. 
For example, generating future tokens~\cite{radford2018improving}, predicting masked tokens \cite{devlin2019bert}, and denoising corrupted tokens~\cite{lewis2020bart} are common pretext tasks in the area of natural language. 
For computer vision, some works use pretexts such as colorization~\cite{zhang2016colorful}, rotation prediction~\cite{gidaris2018unsupervised}, and patch position prediction~\cite{doersch2015unsupervised, noroozi2016unsupervised} to learn representative features for image data.

Most recently, contrastive self-supervised learning has shown great potential by leveraging both negative and positive samples, such as~InstDis~\cite{wu2018unsupervised}, contrastive predictive coding (CPC)~\cite{oord2018representation}, AMDIM~\cite{bachman2019learning}, MoCo~\cite{he2020momentum, chen2020improved}, SimCLR~\cite{chen2020simple, chen2020big}, and InfoMin~\cite{tian2020makes}. 
These methods usually pull together two augmented views of the same object (positive samples) to encourage local invariance while pushing apart those of different objects (negative samples).
This strategy could prevent the model from mapping all instances to the same representation, \ie~representational collapse. However, the contrastive pairs are not easy to be appropriately constructed and need a large batch size or memory bank for storage~\cite{tian2021understanding}. 
To solve this problem, some non-contrastive approaches are developed with only using positive pairs but achieving remarkable performance, such as BYOL~\cite{grill2020bootstrap}, SwAV~\cite{caron2020unsupervised}, SimSiam~\cite{chen2021exploring}, DirectPred~\cite{tian2021understanding}, and DINO~\cite{caron2021emerging}.
The non-contrastive methods use a siamese-like network to match two augmented views of the same object. 
Typically, one network is updated online, and another is directly constructed using the online one. 
Without contrasting negative instances, the training process of non-contrastive methods is more efficient and conceptually simple~\cite{chen2021exploring, tian2021understanding}.

With this in mind, we choose the non-contrastive direction and specify appropriate augmentations for different types of data domains to enhance the instance discrimination in our proposed method.

\subsection{Transfer Clustering}
The NCD task can also be considered as a ``transfer clustering'' problem in \cite{han2019learning, han2019automatically, han2021autonovel}. 
Different from the classic or deep learning clustering problem, transfer clustering first learns the appropriate criterion by using the labelled data, then transfers such knowledge to partition the unlabelled data with novel categories. 
Specifically, Hsu \etal ~\cite{hsu2018learning} propose to learn the category-agnostic pairwise similarity with the Kullback–Leibler divergence based contrastive loss (KCL) on the labelled data. 
This semantic information is then transferred to the unlabelled data by training the model with binary pseudo labels. The~\cite{hsu2018multi} improves KCL with a novel meta classification likelihood (MCL) loss. 
In~\cite{han2019learning}, the deep embedded clustering (DEC)~\cite{xie2016unsupervised} is extended to conduct joint transfer clustering and representation learning.
On the other hand, AutoNovel~\cite{han2019automatically, han2021autonovel} utilizes the pretext task of rotation predictions for self-supervised learning and transfers knowledge of labelled classes to the clustering of unlabelled data by using ranking statistics. 
Jia \etal ~\cite{jia2021joint} extend the noise-contrastive estimation in self-supervised representation learning to jointly handle labelled and unlabelled data. 

In contrast to existing works that mainly use instance discrimination to help separate the novel visual concepts, we show that both instance and category discrimination are essential. 
To the best of our knowledge, we are the first to focus more on category discrimination and creatively make use of prototypes to alleviate the issues with missing labels for the NCD task. 

\section{Proposed Method}
\label{S3}
Given some labelled images from \textit{base} categories, the goal of novel category discovery is automatically discovering \textit{novel} categories in the test image collection~\cite{han2019learning, han2019automatically}. 
In the training stage, we have access to the labelled data $D^l$ and the unlabelled data $D^u$. 
The images in $D^l$ are annotated by a set of base categories $C^l$. $D^l=\{(\mathrm{\mathbf{x}}^{l}_{i}, y^{l}_{i}),i=1,...,N^l\}$, where $y^{l}_{i}$ is the corresponding class label for image $\mathrm{\mathbf{x}}^{l}_{i}$ and $N^l$ is the number of labelled data. 
For the images in unlabelled data $D^u$, they belong to the novel categories $C^u$. 
We are only aware of the number of novel categories but not the concrete meaning of each one. 
Note that the base and novel categories are \textit{disjoint}, \ie~$C^{l}\cap C^{u}=\varnothing$. 
In the inference stage, we should assign one of all categories $C^{l}\cup C^{u}$ to each image in the test split. 
Overall, the NCD task targets to transfer the knowledge learned from the labelled data $D^l$ and the unlabelled one $D^u$ to recognize novel categories $C^u$. 
This process is similar to human beings, where we could automatically differentiate some new concepts from the learned knowledge in the past.

Similar to existing schemes, our method mainly consists of two stages to solve the NCD problem, including forming a robust feature extractor and an accurate classifier. 
The former stage obtains effective features of images in both base and novel categories, whereas the latter aims to precise recognition. 
Instead of using the binary similarity~\cite{hsu2018multi, hsu2018learning, han2019learning, han2019automatically, han2021autonovel} in other NCD works, we are the first to exploit prototypes to enhance both stages. 
This prototype could help our model obtain statistics across each category whatever the base or novel. 
For brevity, we name our two stages prototypical representation learning and prototypical self-training. 

\subsection{Stage I: Prototypical Representation Learning}
\label{sec3.1}
The goal of stage I is to obtain a robust feature extractor serving all images, irrespective of the corresponding categories (base or novel). 
The extracted feature should perform well in both instance and category discrimination. 
We adopt the DINO model and prototype learning to improve discrimination abilities.
The loss function $L_{\text{s1}}$ of stage I is:
\begin{align}
L_{\text{s1}} = L_{\text{ins}} + L_{\text{cat}},
\end{align}
where the $L_{\text{ins}}$ and $L_{\text{cat}}$ are the losses for instance and category discrimination.

\subsubsection{Instance Discrimination}
\label{3.1.1}
We boost the model ability in instance discrimination by self-supervised representation learning. 
In this work, we train a model using the self-distillation with no labels (DINO)~\cite{caron2021emerging} scheme on labelled and unlabelled data with uniform sampling by exploiting two properties.
First, the DINO model could be used as the nearest neighbour classifier for the non-parametric clustering. 
This property is consistent with the NCD task, which also requires clustering. 
Second, the DINO method converges fast by self-distillation in the training period. 
The parameters of the teacher are momentum updated by weighted averaging several student models in different training iterations. 
This procedure is like the popular AdaBoost, which obtains the most convincing results by voting various weak classifiers. 
Therefore, the DINO model could present a high-quality instance representation even without annotations.

In contrast to the original DINO scheme, we propose the \textit{restricted rotation}, a new data augmentation strategy, to satisfy different data types. 
In DINO or other self-supervised approaches~\cite{chen2020simple, misra2020self, caron2020unsupervised}, random cropping is widely utilized to extract intrinsic information of the image since it establishes a part-based invariance assumption which is valid for natural object-centric images like ImageNet.
However, this assumption is not reasonable for symbolic data like the OmniGlot. 
Compared to nature images, the symbolic object has a minor appearance like the texture and color. 
Thus it should require a more abstract understanding of self-supervised models. 
In our experiments, we find that random cropping will destroy the structural information of symbolic data, leading to a dramatic performance decrease, see Sec. \ref{4.4.1}. 
To tackle this problem, we design an augmentation strategy named \textit{restricted rotation}, which constructs different random rotated views for a given image in a restricted degree $\theta$. 
The proposed approach injects randomness while keeping the semantic information well for the symbolic data.

We construct an image set $V$ for each input image $x$. 
The set $V$ contains two augmented global views $x_1^g$ and $x_2^g$ without any rotation, and several rotated and local augmented views $x^{\prime}$.
Then we encourage the model to learn the ``local-to-global" \cite{caron2021emerging} and ``rotation-to-upright" correspondences from the image itself by minimizing the following loss:
\begin{align}
L_{\text{ins}} = \frac{1}{N}\sum_{x\in\{x_1^g,x_2^g\}}~\sum_{\substack{x^{\prime}\in V\\x^{\prime}\neq x}}~H(P_{t}(x),P_{s}(x^{\prime})),
\end{align}
where $N$ is the number of samples in a batch, $P$ is the output distribution of the DINO head, and $H(a,b)=-a\log b$.

With the above essential representation learning, a strong baseline for NCD is established (see Sec. \ref{4.4.1}).
Meanwhile, it can facilitate the category discrimination. %we will describe next.

\subsubsection{Category Discrimination}
\label{sec3.1.2}
Inspired by semi-supervised learning on classification or detection, the pseudo-training signals on unlabelled data could improve the recognition performance of the models due to the consistency with labelled data. 
Thus we aim to build a unified classifier for both base and novel categories on labelled and unlabelled data. 
For labelled data $D^l$, we can easily use cross-entropy loss to encode category information directly. 
However, this encoding process is impractical for the unlabelled data $D^u$ without knowing classes. 
As such, we leverage prototype learning to encode the information of novel classes implicitly, thereby generating the pseudo labels online. 
In this way, the labelled and unlabelled data in the same label system would be used in training together.

\subsubsubsection{Online Prototype Learning}
The online prototype learning (OPL) module is proposed to provide online pseudo labels for unlabelled data $D^u$, which can be used to train a cross-entropy criterion simultaneously with labelled data $D^l$. %The online prototype learning (OPL) module on $D^u$ is proposed to be simultaneous with labelled data $D^l$.
Specifically, the OPL constructs a feature prototype for each novel category and generates pseudo labels for discriminative training in an online manner. 
During training, OPL contains the following three steps:

\begin{figure}[!t]
\centering
\includegraphics[width=3.4in]{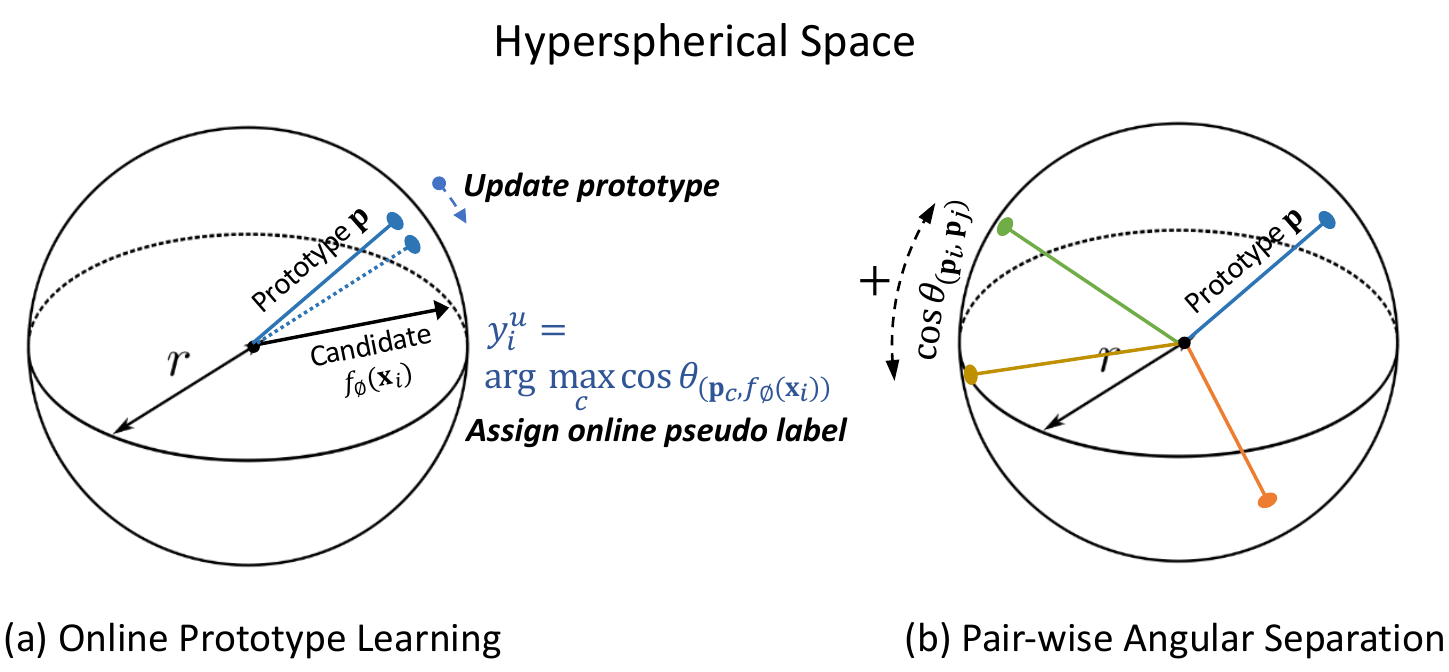}
\caption{Illustration of the online prototype learning (OPL) and pair-wise angular separation (PAS) in stage I. The prototype $\mathrm{\mathbf{p}}$ and candidate feature $f_{\phi}(\mathrm{\mathbf{x}}_{i})$ are re-projected onto the hypersphere. 
During training, OPL provides online pseudo labels for the category discrimination and gradually updates the prototype with its corresponding candidate feature.
Meanwhile, PAS pushes the prototypes far away from each other based on the angular similarity to guarantee effective discrimination.
}
\label{figure_catdis}
\end{figure}

\noindent \textbf{(1) Initializing the prototypes.} At the beginning, there are no well-established classification heads and prototypes for novel categories which need some initialization. 
Specifically, the classification heads of novel categories are initialized with a uniform distribution weight $\mathrm{\mathbf{w}}_{c}$ and no bias value $\mathrm{\mathbf{b}}_{c}=0$ where $c$ denotes the $c$-th category. 
For ease of measuring the cosine similarity, class prototypes are initialized as the L2-normed classifier weights:
\begin{align}
\mathrm{\mathbf{p}}_{c}^{init} = \mathrm{\mathbf{w}}_{c}^{init}/\Vert \mathrm{\mathbf{w}}_{c}^{init}\Vert_{2}.
\label{eq_init}
\end{align}

\noindent \textbf{(2) Assigning online pseudo labels.} For each batch, we calculate the cosine similarity between the $c$-th class prototype $\mathrm{\mathbf{p}}_{c}$ and features of the $i$-th unlabelled data $f_{\phi}(\mathrm{\mathbf{x}}_{i})$ to assign pseudo labels in an online manner:
\begin{align}
y_{i}^u= \arg\max_{c}~
\cos{\theta_{(\mathrm{\mathbf{p}}_{c}, f_{\phi}(\mathrm{\mathbf{x}}_{i}))}}. 
\end{align}
Then, the pseudo labels $y^{u}_{i}$ in the novel category are included to train a classifier responsible for both base and novel categories. We minimize the standard cross-entropy loss for classification: 
\begin{equation}
L_{cls}= -\frac{1}{N}\sum_{i=0}^{N-1} \sum_{c=0}^{C-1} p_{c,i}\log(y_{c,i}),
\label{eq5}
\end{equation}
where $C$ is the number of all categories in $C^l\cup C^u$, and $p_{c,i}$ is the probability of the $i$-th sample for the $c$-th class, $y_{c,i}$ is a binary value that denotes if the $i$-th example belongs to the $c$-th class.

\noindent \textbf{(3) Updating online prototypes.} After assigning the pseudo labels to current batch, we can update the corresponding prototypes using the output features $f_{\phi}(\mathrm{\mathbf{x}}_{i})$.
In particular, the prototype $\mathrm{\mathbf{p}}_{c}$ for the novel category $c$ is updated using the exponential moving average: 
\begin{equation}
\mathrm{\mathbf{p}}_{c} \leftarrow \beta \cdot \mathrm{\mathbf{p}}_{c}+(1-\beta)\cdot f_{\phi}(\mathrm{\mathbf{x}}_{i}), \text{~s.t.~} \  \mathrm{\mathbf{x}}_{i} \in D^u,
\end{equation}
where $\beta\in[0,1)$ is a rate parameter. 
As the model and its assignments become better, the old error-prone features fade, and recent data batch gradually arrives and plays a more important role. 
However, the norm of the prototype may no longer be unit with this modification. 
To facilitate the calculation of cosine similarity, we re-project the prototype onto a hypersphere with the L2-normalization (as in Eq. \ref{eq_init}) after the update.

\noindent \textbf{Avoid trivial solutions.} 
The methods that jointly learn a discriminative classifier and assign pseudo labels would suffer from the problem of trivial solutions~\cite{caron2018deep}. 
A similar phenomenon occurs in directly training our model with OPL. 
The assignments are collapsed into a single prototype, thereby leading to the classifier predicting a single class. 
Considering the situation, we use a uniform class distribution for initial pseudo labels at the first iteration of the training epoch. 

\noindent \textbf{Pair-wise Angular Separation.}
With the help of the L2-normalization (Eq.~\ref{eq_init}) in the prototype initialization and update, we could project prototypes onto a hypersphere.
In this non-Euclidean output space, the distance is evaluated by the \textit{angular} similarity (\ie~cosine similarity) between outputs and class prototypes. 
Recall the goal of obtaining a discriminative feature extractor in stage I. 
We encourage class prototypes to have significant angular separation during prototype learning. 
This idea is intuitive and effective \cite{mettes2019hyperspherical}: the distance between two semantically-unrelated classes would be pushed away if their corresponding class prototypes were positioned separately on the hypersphere.

Since there is no optimal separation algorithm for three- or higher-dimensional unit-hypersphere (known as the Tammes problem~\cite{tammes1930origin}), we optionally approximate the separation by maximizing the cosine distances of prototypes. 
Following \cite{mettes2019hyperspherical}, we define a cosine similarities loss over each pairwise prototypes: 
\begin{align}
    L_{pas} = \frac{1}{K}\sum_{i=1}^{K}\max_{j\in C^u} \mathrm{\mathbf{M}}_{i,j},~~ \mathrm{\mathbf{M}}=\mathrm{\mathbf{P}}\mathrm{\mathbf{P}}^{\top}
-2\mathrm{\mathbf{I}}
\end{align}
\iffalse
\begin{align}
    \mathrm{\mathbf{P}}^* = \mathop{\arg\min}_{\mathrm{\mathbf{P}}^{\prime}\in\mathbb{P}}\left(\max_{(i,j,i\neq j)\in C^u}\cos{\theta_{(\mathrm{\mathbf{p}}^{\prime}_{i}, \mathrm{\mathbf{p}}^{\prime}_{j})}}\right),
\end{align}
\fi
where $\mathrm{\mathbf{P}}\in \mathbb{R}^{K\times D}$ is the matrix of prototypes, $\mathrm{\mathbf{I}}$ denotes the identity matrix in case of self-selection, and $\mathrm{\mathbf{M}}$ is the final pairwise prototypes similarities. 
Different from \cite{mettes2019hyperspherical} which defines class prototypes \textit{a priori} with large margin separation, we use data-dependent class prototypes that are updated by instreaming novel instances. 
Hence the loss function can be simultaneously optimized with the learning process in stage I.

\subsubsubsection{Joint Optimization}
Finally, the category discrimination loss $L_{cat}$ is computed by:
\begin{align}
     L_{cat} = L_{cls} + \lambda L_{pas},
\label{eq_cat}
\end{align}
where the $L_{cls}$ is the cross-entropy loss for classification on both $D^l$ with human labels and $D^u$ with pseudo labels generated from the prototype learning; and $L_{pas}$ is the loss for pair-wise angular separation.

\subsection{Stage \uppercase\expandafter{\romannumeral2}: Prototypical Self-training}
\label{sec3.3}
While a robust feature extractor has been learned in stage \uppercase\expandafter{\romannumeral1}, we empirically find the online pseudo labels are of less quality than offline ones (see Sec. \ref{4.4.2}). 
Therefore, in this stage, we discard the online classifier of stage \uppercase\expandafter{\romannumeral1} and retrain a parametric classifier that recognizes both base and novel categories. 
Specifically, we utilize offline pseudo labels to conduct prototypical self-training. 
The reasons for using self-training in NCD are two-fold: (1) we do not have annotations of novel data $D^u$, while we can generate offline pseudo labels via non-parametric recognition based on stage \uppercase\expandafter{\romannumeral1}; (2) recent approaches on classification, detection or segmentation~\cite{xie2020self,zoph2020rethinking} achieve improvements in self-training even after using popular self-supervised pre-training and supervised learning.

Based on the simple yet effective self-training methods~\cite{xie2020self, lee2013pseudo}, our prototypical self-training includes three steps. 
First, we use the model trained in stage \uppercase\expandafter{\romannumeral1} to generate pseudo labels on unlabelled data (Sec.~\ref{sec3.2.1}).
Then, we note that the loss for pseudo labels would be rectified based on class prototypes (Sec.~\ref{sec3.2.2}).
Finally, we retrain the model by optimizing the classification loss on both human and offline pseudo labels (Sec.~\ref{sec3.2.3}).

\subsubsection{Pseudo labelling}
\label{sec3.2.1}
Based on the well-trained feature extractor in stage~\uppercase\expandafter{\romannumeral1}, we collect all novel images and use the $k$-means~\cite{macqueen1967some} clustering method to generate offline pseudo labels at the start of stage~\uppercase\expandafter{\romannumeral2}. 
In the pseudo labelling on symbolic data, we observe features typically in the non-flat regions, such as the alphabets in OmniGlot~\cite{lake2015human}. 
%
%MH: symbolic images? what do you mean?
Hence we choose the spectral clustering~\cite{ng2001spectral} to separate symbolic data. 
After clustering, we obtain cluster labels as offline pseudo labels for all novel images $D^{u}$. 
In addition, the central feature of clusters is re-positioned as new class prototypes.

\subsubsection{Prototypical Pseudo Label Rectification}
\label{sec3.2.2}
Next, we use prototypes and angular/cosine similarity to rectify the self-training from noisy pseudo labels on $D^u$. 
Instead of the hard filtering, \ie~only images whose prediction confidence of pseudo label higher than a given threshold are considered in the training, we use the soft weighting for label rectification. 
We formulate the rectified objective with the angular/cosine similarity as:
\begin{align}
    L_{rect} = -\sum_{i=0}^{N^u-1} \cos{\theta_{(\mathrm{\mathbf{p}}_{c}, f_{\phi}(\mathrm{\mathbf{x}}_{i}))}} \sum_{c=0}^{C-1} p_{c,i}\log(y_{c,i}), \\ 
    \text{~s.t.~} \mathrm{\mathbf{x}}_{i} \in D^u. \nonumber
\label{l_rect}
\end{align}
Note that the class prototype $\mathrm{\mathbf{p}}_{c}$ is fixed during the training epoch.

\subsubsection{Joint Optimization}
\label{sec3.2.3}
We jointly optimize the loss of both base data with human labels and novel data with offline pseudo labels. For the base data, we utilize the standard cross-entropy loss:
\begin{align}
    L_{ce} = -\sum_{i=0}^{N^l-1} \sum_{c=0}^{C-1} p_{c,i}\log(y_{c,i}), \text{~s.t.~} \mathrm{\mathbf{x}}_{i} \in D^l.
\end{align}
This loss is similar to Eq.~\ref{eq5}, but here we only use instances in the labelled data $D^l$. %
Together with the rectified loss $L_{rect}$ in Eq.~\ref{l_rect} that uses novel categories in the unlabelled data $D^u$, the overall objective is
\begin{align}
    L_{s2}=\frac{1}{N^u+N^l}(L_{ce} + L_{rect}).
\end{align}
After joint optimization, we obtain the enhanced model with an explicit classification layer. 
This model can be reused to generate new offline pseudo labels. 
That is, the prototypical self-training procedure can optionally iterate to further refine the decision boundary of classifiers.

%------------------------------------------------------------------------

\section{Experiments}
\label{S4}

%%% TABLE_splits
\begin{table}[t]
\caption{Dataset splits for novel category discovery experiments.}
\label{table_splits}
\vspace{-4mm}
\small
\begin{center}
\linespread{1.2}\selectfont
\begin{tabular}{ccc}
\toprule 
\multirow{1}*{~~~~~~Dataset~~~~~~} &$\#$ labelled cls& $\#$ unlabelled cls \\
\hline
\multirow{1}*{OmniGlot}&964&659\\
\multirow{1}*{CIFAR10}&5&5\\
\multirow{1}*{CIFAR100}&80&20\\
\multirow{1}*{ImageNet}&882&118\\
\bottomrule
\end{tabular}
\end{center}
\end{table}

\subsection{Datasets and Implementation Details}
\subsubsection{Datasets}
The proposed method is extensively evaluated on four benchmark datasets: CIFAR10 \cite{krizhevsky2009learning}, CIFAR100 \cite{krizhevsky2009learning}, OmniGlot \cite{lake2015human}, and ImageNet \cite{deng2009imagenet}. 
Following previous works \cite{han2019automatically, han2021autonovel}, the dataset splits for novel category discovery experiments are shown in Table \ref{table_splits}. 
Next, we briefly introduce the datasets and describe experimental setups. 

\noindent \textbf{CIFAR10 and CIFAR100.} 
There are 10 object classes in CIFAR10, and each class has 5,000 training and 1,000 testing images of $32\times 32$ resolution. 
Following  \cite{han2019automatically, han2021autonovel}, CIFAR-10 is separated into labelled and unlabelled subsets. 
The first 5 categories (\ie~airplane, automobile, bird, cat, deer) are the labelled set, and the last 5 categories (\ie~dog, frog, horse, ship, truck) are the unlabelled set. 
CIFAR100 also contains 50,000 training and 10,000 testing images, but with a total of 100 classes. 
Each class includes 500 training and 100 testing images in $32\times 32$ resolution. 
For NCD, the first 80 classes are selected as labelled data, and the remaining 20 classes are used as unlabelled data.

\noindent \textbf{OmniGlot.} OmniGlot is a challenging dataset of handwritten characters. 
It contains a total of 1,623 characters from 50 different alphabets, and each alphabet has 20$\sim$47 characters. 
OmniGlot splits 30 alphabets as the ``background'' set and 20 alphabets as the ``evaluation'' set. 
Our experimental setting for NCD follows \cite{hsu2018multi} and \cite{han2019automatically}. 
Specifically, the 30 alphabets in the ``background'' are set as labelled data, including 969 characters (classes). 
The 20 alphabets in the ``evaluation'' are set as unlabelled data, which contain 659 characters. 
The results of the OmniGlot dataset are averaged across the 20 alphabets in the ``evaluation'' set.

\noindent \textbf{ImageNet.} ImageNet is a large-scale visual dataset that contains 1,000 classes with about 1,000 images per class. 
As in \cite{vinyals2016matching}, the ImageNet dataset is randomly split into the 882-class and 118-class subsets. 
Following previous works \cite{hsu2018multi, han2019learning, han2019automatically, han2021autonovel}, we use the 882-class ImageNet as labelled data, then use three 30-class subsets ($\sim$39k images each subset) randomly sampled from the 118-class ImageNet as unlabelled sets. 
As in \cite{han2019learning, han2021autonovel}, the results are averaged over three 30-class subsets.

\begin{figure*}[t]
\centering
\includegraphics[width=5.95in]{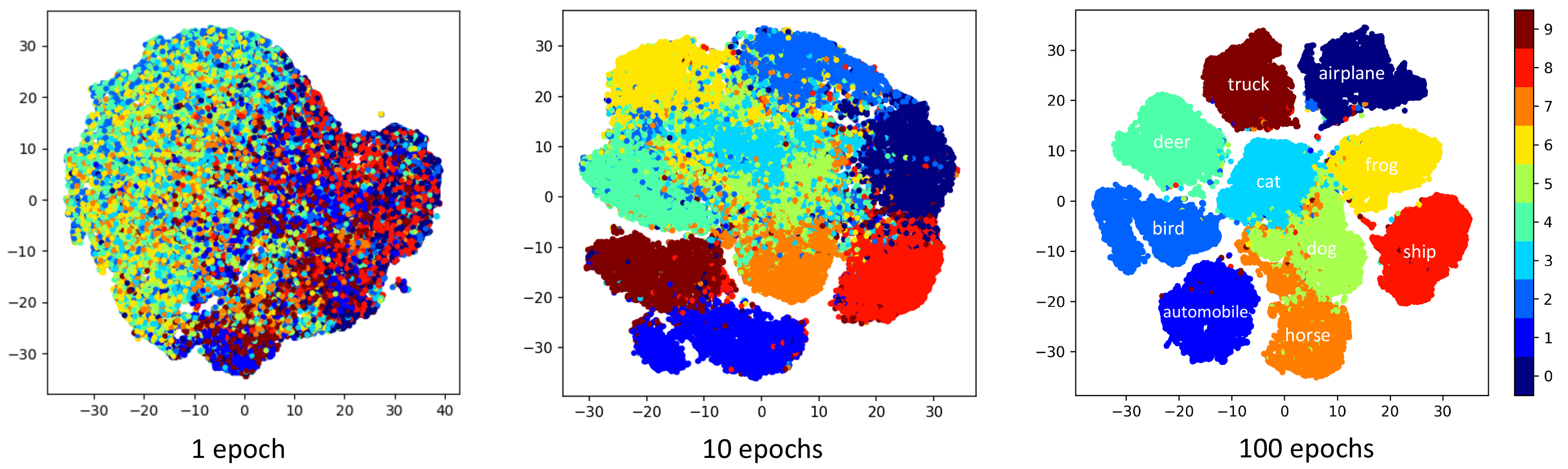}
\caption{Visualized features of unlabelled data in CIFAR10 using the t-SNE \cite{van2008visualizing} projection. The 1, 10, and 100 epochs denote the evolution during the training phase. Colors for class numbers 5-9 refer to ground truths of five novel categories: dog, frog, horse, ship, and truck.
}
\label{figure_qualitative}
\end{figure*}

\subsubsection{Implementation Details}
\label{4.1.2}
Following previous works \cite{han2019learning, han2019automatically, han2021autonovel, jia2021joint}, we use the \textit{clustering accuracy} (ACC) that denotes the matching accuracy between ground-truth labels and clustering assignments to evaluate the performance of our method. 
The results are averaged over 10 runs for all datasets except the ImageNet. Our method is implemented with PyTorch 1.7.1 and runs on the NVIDIA 2080Ti GPUs. 

In stage I, our method jointly performs instance and category discrimination on the labelled training data $D^l$ and unlabelled training data $D^{u}$ with online pseudo labels. 
Specifically, we train our model for 100 epochs on the CIFAR10, CIFAR100, OmniGlot, and ImageNet datasets with the AdamW optimizer \cite{loshchilov2018fixing}. 
During the first 10 epochs, the learning rate is linearly warmed up to the base value determined with the linear scaling rule: $lr = 0.0005*\rm{batchsize}/256$, in which we set $\rm{batchsize}=256$. 
After that, the learning rate is decayed with a cosine schedule~\cite{loshchilov2016sgdr, caron2021emerging}. 
For fair comparisons to previous works, we use ResNet-18 \cite{he2016deep} as our backbone.
In Eq. \ref{eq_cat}, we set $\lambda=0.1$ for all datasets. 
The embedding dimension of prototypes is set to 512.
The data augmentation strategies for different domains are described in detail in Section \ref{4.4.1}. 

For stage II, we use two iterations (two epochs for each iteration) of the self-training for all datasets to further improve the performance. 
The learning rate is set to 0.05 and decayed with a cosine schedule as the same in stage I. 
%

%%% TABLE_results1_cifar
\begin{table}[]
\caption{Comparative performance for the novel category discovery on the \textit{unlabelled training set} of CIFAR10 and CIFAR100. ``w/ S.S.'' means with self-supervision, ``w/ I.L.'' denotes with incremental learning.}
\label{table_results1_cifar}
\vspace{-4mm}
\small
\begin{center}
\linespread{1.15}\selectfont
\begin{tabular}{ccc}
\toprule  
\multirow{1}*{~~Method~~}&~~CIFAR10~~&~~CIFAR100~~ \\
\hline
\multirow{1}*{$k$-means \cite{macqueen1967some}}&65.5$\pm$0.0\%&56.6$\pm$1.6\%\\
\multirow{1}*{KCL \cite{hsu2018learning}}&66.5$\pm$3.9\%&14.3$\pm$1.3\%\\
\multirow{1}*{MCL \cite{hsu2018multi}}&64.2$\pm$0.1\%&21.3$\pm$3.4\%\\
\multirow{1}*{DTC \cite{han2019learning}}&87.5$\pm$0.3\%&56.7$\pm$1.2\%\\
\multirow{1}*{DTC \cite{han2019learning} w/ S.S. \cite{han2019automatically}}&88.7$\pm$0.3\%&67.3$\pm$1.2\%\\
\multirow{1}*{AutoNovel \cite{han2021autonovel}}&90.4$\pm$0.5\%&73.2$\pm$2.1\%\\
\multirow{1}*{AutoNovel \cite{han2021autonovel} w/ I.L.}&91.7$\pm$0.9\%&75.2$\pm$4.2\%\\
\multirow{1}*{WTA-NCD~\cite{jia2021joint}}&93.4$\pm$0.6\%&76.4$\pm$2.8\%\\
\hline
\multirow{1}*{Ours}&\textbf{96.0$\pm$0.4\%}&\textbf{78.9$\pm$0.9\%}\\
\bottomrule
\end{tabular}
\end{center}
\end{table}

\begin{table*}[!htbp]
\caption{Comparative performance for recognizing both old and new categories on the \textit{testing set} of CIFAR10 and CIFAR100.}
\label{table_results2_cifar}
\vspace{-4mm}
\small
\begin{center}
\linespread{1.2}\selectfont
\begin{tabular}{cccccccccc}
\toprule 
\multicolumn{4}{c}{\multirow{2}*{Method}} &  \multicolumn{3}{c}{CIFAR10} & \multicolumn{3}{c}{CIFAR100}\\
\cline{5-10}
&&&&old &new &all &old &new  &all \\
\hline
\multicolumn{4}{c}{KCL \cite{hsu2018learning} w/ S.S.}&~79.4$\pm$0.6\%~ &~60.1$\pm$0.6\%~ &~69.8$\pm$0.1\%~ &~23.4$\pm$0.3\%~ &~29.4$\pm$0.3\%~ &~24.6$\pm$0.2\%~\\
\multicolumn{4}{c}{MCL \cite{hsu2018multi} w/ S.S.}&~81.4$\pm$0.4\%~ &~64.8$\pm$0.4\%~ &~73.1$\pm$0.1\%~ &~18.2$\pm$0.3\%~ &~18.0$\pm$0.1\%~ &~18.2$\pm$0.2\%~\\
\multicolumn{4}{c}{DTC \cite{han2019learning} w/ S.S.}&~58.7$\pm$0.6\%~ &~78.6$\pm$0.2\%~ &~68.7$\pm$0.3\%~ &~47.6$\pm$0.2\%~ &~49.1$\pm$0.2\%~ &~47.9$\pm$0.2\%~\\
\multicolumn{4}{c}{AutoNovel \cite{han2021autonovel} w/ I.L.}&~90.6$\pm$0.2\%~ &~88.8$\pm$0.2\%~ &~89.7$\pm$0.1\%~ &~71.2$\pm$0.1\%~ &~56.8$\pm$0.3\%~ &~68.3$\pm$0.1\%~\\
\hline
\multicolumn{4}{c}{Ours}&~\textbf{97.1$\pm$0.4\%}~&~\textbf{93.8$\pm$0.5\%}~&~\textbf{95.4$\pm$0.4\%}~&~\textbf{76.9$\pm$0.3\%}~&~\textbf{64.0$\pm$0.6\%}~&~\textbf{74.3$\pm$0.4\%}~\\
\bottomrule
\end{tabular}
\end{center}
\end{table*}

\subsection{Novel Category Discovery}
\subsubsection{Comparison with State-of-the-art Methods}
We first evaluate the proposed method and compare it with other state-of-the-art methods for novel category discovery on the \textit{unlabelled training set} of CIFAR10 and CIFAR100. 
Table \ref{table_results1_cifar} shows that our method achieves 96.0\% and 78.9\% ACC on CIFAR10 and CIFAR100, outperforming the previous works. 
Note that ``$k$-means'', shown in the first row of Table \ref{table_results1_cifar}, is our baseline. 
It represents directly training a model using $D^l$ and applying clustering on novel categories in $D^u$, which only obtains 65.5\% and 56.6\% ACC on CIFAR10 and CIFAR100, respectively. 
We also show evaluation results on the \textit{testing set} of CIFAR10 and CIFAR100 in Table \ref{table_results2_cifar}. 
This experimental setting contains old and new (\ie~base and novel) categories that need to be simultaneously recognized without forgetting. 
As shown in Table \ref{table_results2_cifar}, the proposed method stands out by achieving 97.1\%, 93.6\%, and 95.3\% ACC on the old, new, and all categories of CIFAR10. 
In addiiton, it achieves the best 76.9\%, 63.3\%, and 74.2\% ACC on the same setting in CIFAR100.
Especially, the proposed method shows greater advantages (about 5\%$\sim$7\% ACC improvements) on the testing set than the unlabelled training set, demonstrating its robust generalization ability.

%%% TABLE_results_omni_imgnet
\begin{table}[t]
\caption{Comparative performance for the novel category discovery on Omniglot and ImageNet unlabelled set.}
\label{table_results_omni_imgnet}
\vspace{-4mm}
\small
\begin{center}
\linespread{1.2}\selectfont
\begin{tabular}{ccc}
\toprule  
\multirow{1}*{~~Method~~}&~~OmniGlot~~&~~ImageNet~~ \\
\hline
\multirow{1}*{$k$-means \cite{macqueen1967some}}&77.2\%&71.9\%\\
\multirow{1}*{KCL \cite{hsu2018learning}}&82.4\%&73.8\%\\
\multirow{1}*{MCL \cite{hsu2018multi}}&83.3\%&74.4\%\\
\multirow{1}*{DTC \cite{han2019learning}}&89.0\%&78.3\%\\
\multirow{1}*{AutoNovel \cite{han2021autonovel}}&89.1\%&82.5\%\\
\multirow{1}*{WTA-NCD~\cite{jia2021joint}}&-&86.7\%\\
\hline
\multirow{1}*{Ours }&\textbf{93.4\%}&\textbf{88.8\%}\\
\bottomrule
\end{tabular}
\end{center}
\end{table}

Furthermore, we evaluate our method on the more challenging OmniGlot and ImageNet datasets. 
The results of OmniGlot are averaged over the 20 alphabets in the evaluation set, and the results of ImageNet are average over three random 30-class unlabelled subsets as in previous works \cite{hsu2018multi, han2019learning, han2019automatically, han2021autonovel}. 
As shown in Table \ref{table_results_omni_imgnet}, the proposed method achieves the best performance with 93.4\% and 88.8\% ACC on Omniglot and ImageNet, demonstrating the effectiveness of our approach.

\subsubsection{Qualitative Analysis}
Next, we qualitatively analyze the proposed method for novel category discovery. 
Figure \ref{figure_qualitative} illustrates the evolution of instance features of base (0-4) and novel (5-9) categories in CIFAR10 using the t-SNE \cite{van2008visualizing}. 
The instance features of both base and novel categories become more separable and gradually gather into clusters. 
Meanwhile, some instances of the dog (5) and horse (7) are close since they maintain similar appearances as four-legged mammals. 
We note that mining the discrimination of such similar categories is an interesting problem and worthy of further study.

%%% TABLE_ablations
\begin{table}[t]
\caption{Ablation studies of the proposed method. ``InstDis'' and ``CatDis'' stand for instance discrimination and category discrimination in stage I, respectively. ``PST'' refers to the prototypical self-training in stage II. All methods use the same hyperparameters and are evaluated with clustering accuracy (ACC).}
\label{table_ablations}
\vspace{-4mm}
\small
\begin{center}
\linespread{1.2}\selectfont
\begin{tabular}{cccc}
\toprule 
\multirow{1.1}*{Method}&CIFAR10&CIFAR100&OmniGlot  \\
\hline
\iffalse
\multicolumn{1}{l}{\zl{Ours w/o InstDis}} &\zl{91.0\%} &\zl{64.7\%} &\zl{89.5\%}\\
\multicolumn{1}{l}{\zl{Ours w/o CatDis}}&\zl{92.9\%}  &\zl{72.8\%} &\zl{91.2\%}\\
\multicolumn{1}{l}{\zl{Ours w/o PST}} &\zl{93.6\%} &\zl{75.6\%} &\zl{92.0\%}\\
\hline
\multicolumn{1}{l}{\zl{Ours}}&\zl{\textbf{96.0\%}} &\zl{\textbf{78.9\%}} &\zl{\textbf{93.4\%}}\\
\fi
\multicolumn{1}{l}{Ours w/o InstDis} &91.0\% &64.7\% &89.5\%\\
\multicolumn{1}{l}{Ours w/o CatDis}&92.9\%  &72.8\% &91.2\%\\
\multicolumn{1}{l}{Ours w/o PST} &93.6\% &75.6\% &92.0\%\\
\hline
\multicolumn{1}{l}{Ours}&\textbf{96.0\%} &\textbf{78.9\%} &\textbf{93.4\%}\\
\bottomrule
\end{tabular}
\end{center}
\end{table}

\subsection{Ablation Study}
To analyze the contribution of proposed components in two stages, we conduct several ablation studies on the CIFAR10, CIFAR100, and OmniGlot datasets.

\noindent \textbf{Instance Discrimination.}
We compare performances with and without the instance discrimination module (InstDis) to validate its effectiveness for the novel category discovery. 
As shown in Table \ref{table_ablations}, InstDis is a basic component for learning semantic representations of the unlabelled data. 
For the CIFAR10, CIFAR100, and OmniGlot datasets, InstDis significantly boosts the cluster accuracy by 5.0\%, 14.2\%, and 3.9\%, respectively.

\noindent \textbf{Category Discrimination.}
Table~\ref{table_ablations} illustrates the effectiveness of category discrimination (CatDis). 
As a crucial component of the proposed method, CatDis enhances the class-level discrimination of features and consistently improves performance for the novel category discovery, \ie~introducing 3.1\%, 6.1\%, and 2.2\% ACC improvements on CIFAR10, CIFAR100, OmniGlot respectively, demonstrating the effectiveness of online prototype learning.

\noindent \textbf{Prototypical Self-training.}
Then we ablate the training of explicit classifier in prototypical self-training (PST). It can be observed that PST is a good strategy to further boost the model's performance for novel category discovery, consistently improving 1$\sim$3\% (\ie~96.0\% \vs~93.6\%, 78.9\% \vs ~75.6\%, 93.4\% \vs~92.0\%) ACC on all three datasets.

\subsection{Analysis and Discussion}
\subsubsection{Instance Discrimination}
\label{4.4.1}
First, we analyze the performance of other state-of-the-art self-supervised methods for the instance discrimination in our model. 
Then we study how the augmentations used in self-supervised learning affect different image domains, \ie~the natural and symbolic images.

\noindent \textbf{Alternative self-supervised methods.}
As discussed in Section \ref{3.1.1}, DINO~\cite{caron2021emerging} is adopted in our method to enhance the instance discrimination. 
However, other self-supervised learning methods can be inserted into our model. 
Since DINO is a non-contrastive method, we choose other state-of-the-art contrastive methods as alternatives, including SimCLR~\cite{chen2020simple}, MoCo~\cite{he2020momentum}, and MoCo v2~\cite{chen2020improved}. 
In Table \ref{table_ssl}, we first compare the learned features of different self-supervised methods using $k$-means~\cite{macqueen1967some} as the clustering method. 
DINO significantly outperforms other alternatives as it is an effective nearest neighbour classifier without any fine-tuning or linear classifier~\cite{caron2021emerging}. 
This is a desirable property for $k$-means clustering. 
Then, when combined with our method, DINO is further improved by 6.7\% and 9.3\% on CIFAR10 and CIFAR100, respectively. 
In addition, SimCLR and MoCo are also significantly boosted and achieve comparable performance, which validates the effectiveness of our method and its compatibility with other self-supervised learning methods.

%%% TABLE_other_ssl_methods
\begin{table}[t]
\caption{The performance comparison with different self-supervised learning methods for the instance discrimination in NCD. }
\label{table_ssl}
\vspace{-4mm}
\small
\begin{center}
\linespread{1.2}\selectfont
\begin{tabular}{cccc}
\toprule  
\multicolumn{2}{c}{~~Method~~}&CIFAR10&CIFAR100 \\
\hline
\multirow{4}*{$k$-means \cite{macqueen1967some}}&SimCLR \cite{chen2020simple}&85.2\%&49.7\%\\
&MoCo \cite{he2020momentum}&77.4\%&51.2\%\\
&MoCo v2 \cite{chen2020improved}&81.3\%&54.1\%\\
&DINO \cite{caron2021emerging}&\textbf{89.3\%}&\textbf{69.6\%}\\
\hline
\multirow{4}*{Ours}&SimCLR \cite{chen2020simple}&92.1\%&61.8\%\\
&MoCo \cite{he2020momentum}&90.9\%&62.9\%\\
&MoCo v2 \cite{chen2020improved}&93.8\%&65.3\%\\
&DINO \cite{caron2021emerging}&\textbf{96.0\%}&\textbf{78.9\%}\\
\bottomrule
\end{tabular}
\end{center}
\end{table}

\noindent \textbf{Different domains and augmentations.}
%Regarding different domains, 
We consider two main categories of domains \cite{wallace2020extending}, natural (CIFAR10) and symbolic (Omniglot), for experiments. 
To systematically study the impact of data augmentation, we follow SimCLR \cite{chen2020simple} to divide augmentations into two types. 
One type of augmentation involves \textit{geometric/spatial} transformation of data, such as cropping, flipping, and the proposed restricted rotation. 
The other type of augmentation involves \textit{appearance} transformation, such as color distortion (including color jittering and dropping, solarization) and Gaussian blur. 
We report the performance of different compositions of data augmentations for both natural and symbolic data in Table~\ref{table_augmentations_rr}. 
For symbolic data, the proposed restricted rotation and appearance transformation contribute a lot to self-supervised learning. 
In contrast, geometric transformations lead to significant performance degradation, since they would negatively affect the structure of symbolic data. 
On the other hand, natural data benefits from the composition of appearance and geometric transformation, as also noted in \cite{chen2020simple}.
Unlike symbolic data, natural data do not need strict structure preservation, hence we do not apply the restricted rotation to them. 
In Table \ref{table_augmentations}, we show the set of augmentations in our implementation for natural and symbolic datasets, respectively.

%%% table_augmentations
\begin{table}[t]
\caption{The performance comparison with different types of data augmentations in the self-supervised feature learning for both natural and symbolic datasets. Green and blue colors denote the best two results.}
\label{table_augmentations_rr}
\vspace{-4mm}
\small
\begin{center}
\linespread{1.1}\selectfont
\begin{tabular}{cccc}
\toprule  
\multirow{1}*{Dataset}&Appearance&Geometric& ACC (\%)\\
\hline
\multirow{5}*{\makecell{OmniGlot \\(symbolic)}}&\checkmark& &90.2\\
&\checkmark& R.R. &\color{Green} 93.4\\
&&\checkmark & 66.3\\
&\checkmark&\checkmark & 88.4\\
&\checkmark& \checkmark~+ R.R. &\color{blue} 91.3\\
\hline
\multirow{5}*{\makecell{CIFAR10 \\(natural)}}&\checkmark & & 79.5 \\
&\checkmark&R.R.& 81.7\\
&&\checkmark &65.9\\
&\checkmark&\checkmark & \color{Green} 96.0\\
&\checkmark&\checkmark~+ R.R.& \color{blue} 95.4\\
\bottomrule
\end{tabular}
\end{center}
\end{table}

%%% table_augmentations
\begin{table}[t]
\caption{Different sets of augmentations (transformations) in the self-supervised feature learning for the natural and symbolic domains.}
\label{table_augmentations}
\vspace{-4mm}
\small
\begin{center}
\linespread{1.1}\selectfont
\begin{tabular}{ccc}
\toprule  
\multirow{1}*{~~Augmentation for S.S~~}&~~Natural~~&~~Symbolic~~ \\
\hline
\linespread{0.95}\selectfont
\multirow{2}*{\tabincell{c}{Brightness, contrast, hue,\\ and saturation adjustment}}&\multirow{2}*{\checkmark} &\multirow{2}*{\checkmark}\\
&&\\
\linespread{1.1}\selectfont
\multirow{1}*{Random cropping}&\checkmark &$\times$ \\
\multirow{1}*{Random left-right flipping}&\checkmark &$\times$ \\
\multirow{1}*{Restricted rotation}&$\times$ &\checkmark\\
\multirow{1}*{Color jittering and dropping}&\checkmark &\checkmark\\
\multirow{1}*{Gaussian blurring, solarization}&\checkmark &\checkmark\\
\bottomrule
\end{tabular}
\end{center}
\end{table}

\subsubsection{Category Discrimination}
\label{4.4.2}
As online prototype learning plays a crucial role in the category discrimination of stage I, we analyze the hyperparameters of prototypes and compare pseudo labels provided by prototype assignment and $k$-means. 

\begin{figure}[!t]
\centering
\includegraphics[width=2.65in]{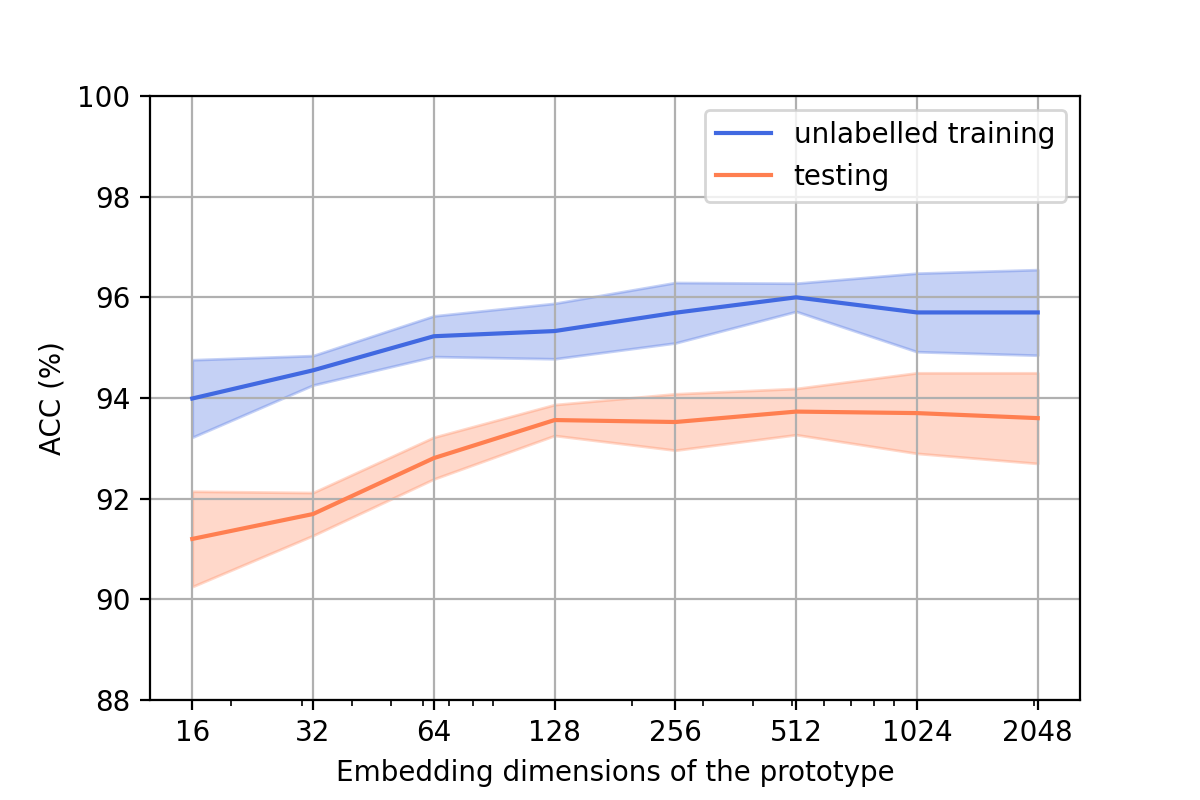}
\caption{Illustration of the performance evolution for different embedding dimensions of the prototype. Experiments are performed on CIFAR10.
}
\label{figure_exp_prototype}
\end{figure}

\noindent \textbf{Hyperparamters of prototypes.} In the online prototype learning 
module, a representative feature of novel categories is maintained as the prototype. 
First, we study how the embedding dimensions of prototypes affect the performance on CIFAR10. 
As shown in Figure \ref{figure_exp_prototype}, the proposed method is stable with different prototype dimensions, and the performances plateau when dimensions are greater than 128. 
Next, we show different similarity metrics for the prototype when assigning online pseudo labels. 
Compared to the Euclidean distance and dot product, our cosine similarity metric achieves the best accuracy on both natural and symbolic datasets, as shown in Table \ref{table_proto_similarity}.

\noindent \textbf{Prototype assignment and $k$-means.} 
As two potential ways to generate pseudo labels in stage I, we compare the prototype assignment and $k$-means in Table \ref{table_pseudo_labels}. 
With the proposed OPL, the results of $k$-means become more accurate. 
Actually, $k$-means is a simple but effective clustering method to evaluate the distinctiveness of our feature extractor in stage I. 
Table \ref{table_pseudo_labels} also shows that the ``online (prototype assignment)'' and ``offline ($k$-mean, ours full)'' have comparable accuracy, and the latter slightly outperforms the former at the end of training. 
This can be attributed to that offline pseudo labels could see the whole dataset, whereas online pseudo labels are generated batch by batch. 
Therefore, we use offline pseudo labels provided by ``$k$-means, ours full'' as the bridge between stages I and II. 
In addition, we note that only online (prototype-based) pseudo labels are used to train the discriminative feature extractor in stage I. 
This is motivated by two factors: (1) online pseudo labels are in a more efficient batch-by-batch way, while $k$-means need to see the whole dataset to calculate the distance between every two data instances and conduct an iterative EM-like process, which is relatively time-consuming; and (2) the separation constraint of online prototypes, \ie~pair-wise angular separation, can also contribute to learning discriminative representations.

%%% TABLE_proto_similarity
\begin{table}[t]
\caption{The performance comparison with different similarity metrics for the prototype when assigning online pseudo labels.}
\label{table_proto_similarity}
\vspace{-4mm}
\small
\begin{center}
\linespread{1.0}\selectfont
\begin{tabular}{ccc}
\toprule 
\multirow{1}*{~~\small{Similarity}~~}&~~\small{CIFAR10}~~&~~\small{OmniGlot}~~ \\
\hline
\multirow{1}*{\small{Euclidean}}&\small{95.4\%}&\small{91.8\%}\\
\multirow{1}*{\small{Dot Product}}&\small{94.7\%} &\small{92.0\%}\\
\multirow{1}*{\small{Cosine (ours)}}&\small{\textbf{96.0\%}}&\small{\textbf{93.4\%}}\\
\bottomrule
\end{tabular}
\end{center}
\end{table}

\begin{table}[!t]
\caption{The performance comparison with online and offline pseudo-labels after stage I on CIFAR10.}
\vspace{-4mm}
\small
\begin{center}
\linespread{1.2}\selectfont
\begin{tabular}{cc}
\toprule  %添加表格头部粗线
\multirow{1}*{~~~Different pseudo-labels at the end of stage I~~~}&~~ACC~~\\
\hline
\multicolumn{1}{l}{~~~Offline ($k$-means, ours full)~~~}&~~\textbf{93.6\%}~~\\
\multicolumn{1}{l}{~~~Offline ($k$-means, ours w/o OPL)~~~}&~~91.1\%~~\\
\multicolumn{1}{l}{~~~Online (prototype assignment)~~~}&~~90.5\%~~\\
\multicolumn{1}{l}{~~~Offline ($k$-means)~~~}&~~65.5\%~~\\
\bottomrule
\end{tabular}
\end{center}
\label{table_pseudo_labels}
\end{table}

\subsubsection{Prototypical Self-training}
Based on the discriminative feature extractor in stage I, we collect all novel images and run the global clustering via the off-the-shelf methods (\eg~$k$-means) to generate offline pseudo labels at the start of stage II (PST). 
Note that the clustering method is conducted only once at the beginning of PST.
During the iteration of PST, offline pseudo labels are provided by the explicit classifier for novel categories. 

\noindent \textbf{The number of iterations.}  We report the performance with different numbers of iterations in Table \ref{table_num_iters}. 
At first, the performance significantly increases with the prototypical self-training, then it plateaus after two iterations. 
Hence we use two iterations of prototypical self-training in our implementation, as described in Section \ref{4.1.2}.
 
\begin{figure}[!t]
\centering
\includegraphics[width=3.6in]{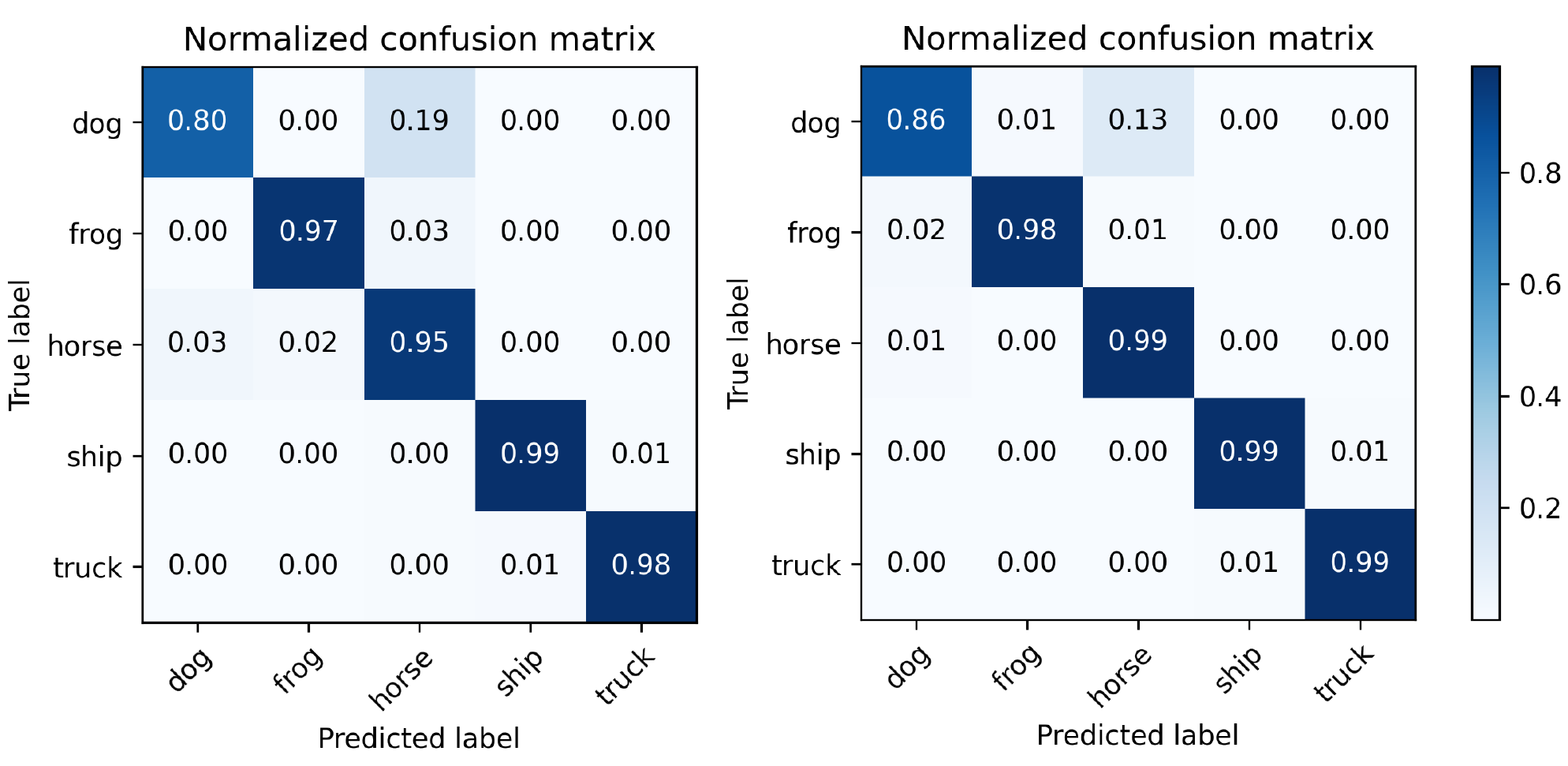}
\caption{Confusion matrix on 5 novel categories of CIFAR10. Left: our method stage I; Right: our method with PST (stage II).
}
\label{figure_confusion_matrix}
\end{figure}

\noindent \textbf{Confusion matrix.} We compare the confusion matrices w/ and w/o PST in Figure \ref{figure_confusion_matrix}. 
The accuracy for novel categories is improved with PST. 
Similar to the t-SNE visualization in Figure \ref{figure_qualitative}, there is confusion between the dog and horse, mainly due to their similar shape and appearance. Therefore, separating similar or fine-grained categories in NCD is of great interest for future study.

\begin{table}[!t]
\caption{The performance evolution for different times of iterations with the PST.}
\label{table_num_iters}
\vspace{-4mm}
\small
\begin{center}
\linespread{1.0}\selectfont
\begin{tabular}{cccccc}
\toprule  
\multirow{1}*{\small{\# Iters}} &\small{0} &\small{1} &\small{2} &\small{3} &\small{4}\\
\hline
\multirow{1}*{\small{CIFAR10}}&\small{93.6\%} &\small{95.6\%} &\small{96.0\%} &\small{\textbf{96.1}\%} &\small{96.0\%}\\
\multirow{1}*{\small{OmniGlot}}&\small{92.0\%} &\small{92.5\%} &\small{\textbf{93.4}\%} & \small{93.3\%} & \small{93.3\%}\\
\bottomrule
\end{tabular}
\end{center}
\end{table}

%------------------------------------------------------------------------
\section{Conclusion}
\label{S5}
%MH: do not use "insight" (big word).
%In this paper, we bring NCD with a novel insight to improve the category discrimination for semantic partition. 
In this paper, we propose a novel method to improve the category discrimination for semantic partition in the NCD task.
Our approach consists of two stages: (1) prototypical representation learning and (2) prototypical self-training. 
Specifically, we leverage the prototype to conduct both instance and category discrimination at stage I, thereby obtaining a robust feature extractor to serve all base and novel images. 
Then, we train a parametric classifier by self-training with prototypical rectified pseudo labels at stage II.  
Extensive experiments on widely-used benchmark datasets show that the proposed method achieves state-of-the-art performance, and demonstrates the effectiveness and robustness of all modules. 
In the future, we plan to explore NCD by including or improving the open-set recognition and incremental learning, and develop a framework for discovering more novel concepts in the real world.

\ifCLASSOPTIONcaptionsoff
  \newpage
\fi

% trigger a \newpage just before the given reference
% number - used to balance the columns on the last page
% adjust value as needed - may need to be readjusted if
% the document is modified later
% \IEEEtriggeratref{8}
% The "triggered" command can be changed if desired:
% \IEEEtriggercmd{\enlargethispage{-5in}}

% references section

% can use a bibliography generated by BibTeX as a .bbl file
% BibTeX documentation can be easily obtained at:
% http://mirror.ctan.org/biblio/bibtex/contrib/doc/
% The IEEEtran BibTeX style support page is at:
% http://www.michaelshell.org/tex/ieeetran/bibtex/
% \bibliographystyle{IEEEtran}
% argument is your BibTeX string definitions and bibliography database(s)
% \bibliography{IEEEabrv,../bib/paper}

% <OR> manually copy in the resultant .bbl file
% set second argument of \begin to the number of references
% (used to reserve space for the reference number labels box)

\bibliographystyle{IEEEtran}
\bibliography{IEEEabrv, mybibfile}

\end{document}